# PRELIMINARY DESIGN OF A DEVICE TO ASSIST HANDWRITING IN CHILDREN WITH MOVEMENT DISORDERS

Gabrielle Lemire[1,2], Thierry Laliberté[1], Katia Turcot[2,3], Véronique Flamand[2,4], Alexandre Campeau-Lecours[1,2]

[1]Department of Mechanical Engineering, Université Laval, Quebec City, Canada, [2]Centre for Interdisciplinary Research in Rehabilitation and Social Integration, CIUSSS de la Capitale-Nationale, Quebec City, Canada, [3]Department of Kinesiology, Université Laval, Quebec City, Canada, [4]Department of Rehabilitation, Université Laval, Quebec City, Canada

## ABSTRACT

This paper presents the development of a new passive assistive handwriting device, which aims to stabilize the motion of people living with movement disorders. Many people living with conditions such as cerebral palsy, stroke, muscular dystrophy or dystonia experience upper limbs impairments (muscle spasticity, unselective motor control, muscle weakness or tremors) and are unable to write or draw on their own. The proposed device is designed to be fixed on a table. A pen is attached to the device using a pen holder, which maintains the pen in a fixed orientation. The user interacts with the device using a handle while mechanical dampers and inertia contribute to the stabilization of the user's movements. The overall mechanical design of the device is first presented, followed by the design of the pen holder mechanism.

## INTRODUCTION

Many people living with conditions such as cerebral palsy, stroke, muscular dystrophy or dystonia experience movement disorders to the upper limbs (muscle spasticity, unselective motor control, muscle weakness or tremors) and are unable to write or draw on their own. Children generally engage in motor activities at school [1]. Taking part in motor activities during childhood foster social inclusion. Those who suffer from motor impairments experience difficulties in learning and participating actively. For them, any school task requires more time, and the results seldom reflect the actual potential of the child. This often leads to social and emotional difficulties [1]. Many school activities, such as drawing, mathematics and composing texts, are based on handwriting. Such activities are difficult or impossible to perform for children living with upper limb impairments. Whereas the use of a computer or a tablet might be an appropriate alternative for children living with theses impairments, it has been demonstrated that handwriting is favorable to the child's development and learning. In comparison with typing, manual writing enhances memorization [2] and letter recognition [3], and is more efficient for note taking [4]. In addition, handwriting helps in the learning process of children in different aspects such as spelling [5] and understanding math problems if and when they write down their approach or calculations [6]. In addition, children learn letters more easily when they write them down, compared to only being shown the letters [7].

Over the years, different assistive technologies (AT) have been developed for people living with upper limb impairments. Some of them have been specifically designed for writing. For instance, ergonomic pens[1] provide a convenient grasp while anti-tremor gloves[2] reduce hand trembling (i.e., for adults living with Parkinson's disease). Wu et al. [8] designed an assistive device supporting the arm of children living with cerebral palsy, to help them draw. Pedemonte et al. [9] created a haptic device to help children reproduce a writing pattern. Shire et al. [10] used the Clinical Kinematic Assessment Tool (CKAT) to teach children how to move correctly in order to write letters. Assistive robotic arms can also be used to write and draw [13, 14]. Assistive devices like those aforementioned aim to help kids with functional impairments in numerous aspects of their lives that require communication and socialization skills [11]. However, following a review of the commercially available products and the scientific literature, a focus group was conducted with rehabilitation researchers and occupational therapists from Quebec, Canada. It was revealed that many children living with various movement disorders, such as dyspraxia, ataxia or spasticity could not write by themselves and that no existing AT was able to meet their needs and help them in that task.

---

[1] Penagain. (n.d.) Retrieved Oct. 2018, from http://penagain.net/
[2] Readi Steadi. (n.d.) Retrieved Oct. 2018, from https://www.readi-steadi.com/

## OBJECTIVES

The objective of this project is to develop a handwriting assistive device giving the ability to control a pen to children living with movement disorders during voluntary (i.e., ataxia) or involuntary movements (i.e., dystonia), and/or upper limb muscular spasticity that leads to abnormal reflex responses, which complicate movement control. The aim of designing a new device is twofold: i) develop an AT that will stabilize the user's motion and enable him/her to draw and write and ii) simplify the design as much as possible for the device to be affordable and universally accessible. The prototype is developed through an iterative process, in collaboration with researchers in engineering and rehabilitation, and occupational therapists, with a user-centered approach based on Design Thinking [12].

## SUMMARY DESCRIPTION

The proposed mechanism, which is designed to be mounted on a table, is shown in Fig. 1. The mechanism has two degrees of freedom (DoF). A pen is attached at the end of the mechanism. The user operates the device by grasping and moving a handle. The orientation of the handle can be adjusted to the user's preference. The device allows moving the pen on the table plane and, as a result of the mechanism design, maintains the handle in a constant orientation. Mechanical inertia and dampers allow stabilization of the user's motion. The device thus assists the user in two different manners: i) the mechanism holds the pen for the user, a task that could prove to be difficult or impossible for some people because of spasticity or upper limb impairments, and ii) inertia and damping stabilize uncoordinated movements (i.e., spasms). Pens of different sizes can be attached to the device thanks to an adaptative pen holder. The sheet is held in place with magnets as the child draws or writes. The mechanical design of the two-DoF mechanism is first presented, followed by the design of the pen holder mechanism.

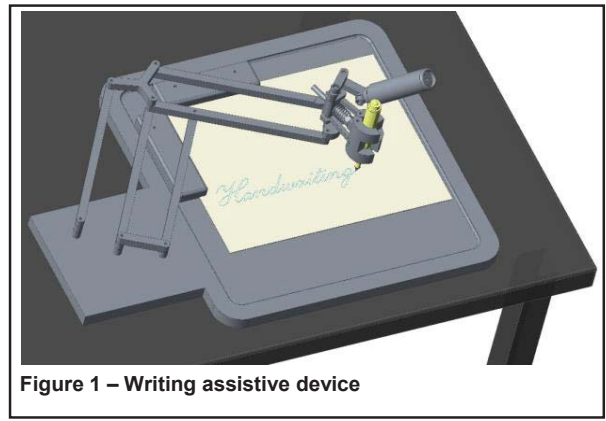

**Figure 1 – Writing assistive device**

## MECHANISM DESIGN

This section presents the mechanism design. Fig. 2 presents three variations of the potential mechanism, in increasing order of complexity, all of which display the same two DoF. The mechanism is inspired by the design of the assistive eating device from Turgeon et al. [15, 16].

Fig. 2a shows a two-DoF mechanism for planar motion using two bars (one for each DoF) and two pivot points (J1 and J2). The mechanism is shown in three different positions. Considering this design, the orientation of the handle, which is attached to the end effector, varies depending on the position of the mechanism. This would require users to adapt to this changing orientation while they draw.

To constrain the orientation of the end-effector relative to the base, two parallelograms are added on each bar, as displayed in Fig. 2b. This allows for the mechanism to still have two DoF, while the orientation of the handle remains constant relative to the base. A triangle links the two parallelograms at J2.

A damper is added at each joint to dampen (J1 and J2). The damper at J2, however, has to be supported by the first links, which should be very stiff in order to compensate for the weight of the damper.

In order to have a mechanism that is lighter and more compact, the second damper is reported to the base, thanks to the use of a third parallelogram as seen in Fig. 2c. Therefore, the two joints (J1 and J2) can be controlled directly from the base of the mechanism. The bars length (25cm) was chosen so that the mechanism covers the entire surface area of a legal size sheet of paper (both portrait and landscape).

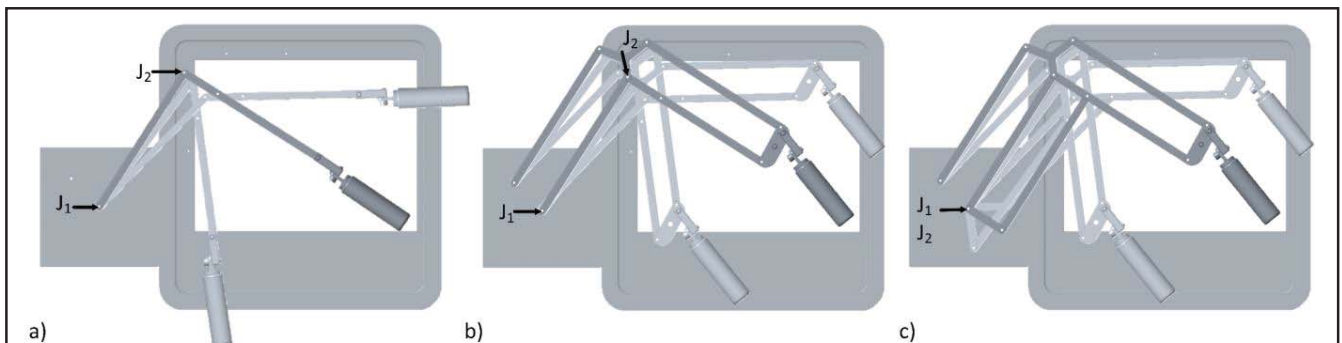

Figure 2 - Development of the mechanism. a) 2-bar mechanism where the end-effector orientation depends on the angles of the two bars. b) 2 parallelograms are added so the end effector stays in the same orientation with respect to the base in all positions. c) A third parallelogram is added for the second rotation to be controlled at the base of the mechanism.

**PEN HOLDER MECHANISM**

The pen is attached at the end of the mechanism and is maintained in a fixed orientation. In the case of a basic design, a simple hole with a set screw would hold the pen in place. However, this would prevent using pens of different sizes. Thus, we aimed to design a mechanism that would firmly hold the pen while being adjustable to many pen sizes. The mechanism can adapt to pen diameters from 8 mm to 20 mm (which is the range of regular pen sizes found on the market). The mechanism that holds the pen can be adjusted quickly and easily, while also providing a tight grip on the pen. The inspiration of the proposed mechanism comes from the grapples designed to grab different sizes of trees. A 2-finger gripper with circular distal members grabs the pen. That motion is initialized with a screw moving on a shaft that activates the proximal members, as shown in Fig. 3a. As these links move, the distal members (i.e., the fingers) close on the pen. The closed mechanism is displayed in Fig. 3b. In Fig. 3c, a small spring is shown, which allows the mechanism to open automatically when the screw is released.

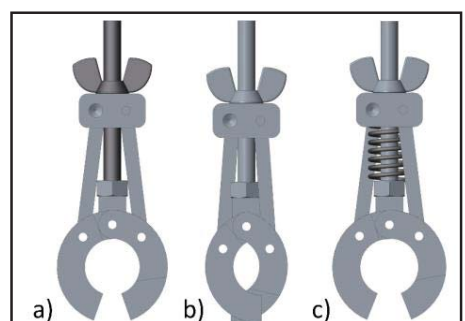

Figure 3 – Pen holder mechanism. a) Dark grey part shows the adjustment mechanism depending on the pen size. b) The mechanism in closed position. c) The spring shown in dark grey is added for the mechanism to automatically open when pressure is released around the pen.

The pen holder mechanism is mounted on the bar at the end of the 2-DoF mechanism, on the last link of the parallelogram, near the handle. The orientation of the pen can be adjusted with a simple screw. It is preferable to mount the handle directly on the pen mechanism since it needs to be adjusted with respect to the position of the pen. Fig. 4 displays the mounting mechanism. The handle is linked to the pen holder using a three-bar mechanism (K1, K2 and K3), which allows three adjustments. The three bars are identified on the figure as K1, K2 and K3. Thus, the position and the orientation of the handle can be adjusted for every user. All the adjustments can be made with simple screws.

The handle can be replaced with custom and other types of handles.

**DISCUSSION**

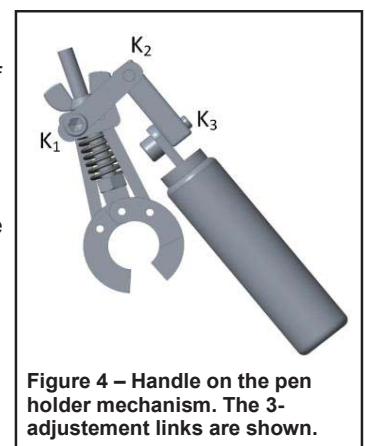

Figure 4 – Handle on the pen holder mechanism. The 3-adjustement links are shown.

In this paper, the prototype of a preliminary handwriting assistive device was presented. The objective is to help children living with movement disorders to write and draw in a learning context. Preliminary discussions with occupational therapists revealed that the device has the potential to help children and adolescents write or draw by themselves. Future work will consist in evaluating the prototype with potential users in order to assess its efficiency in the writing and drawing process.

**CONCLUSION**

This paper has presented a 2-DoF handwriting mechanism designed to assist children living with movement disorders in writing and drawing. The mechanism includes a 2-DoF planar mechanism that keeps the end effector in the same orientation with respect to the base, and a pen holder mechanism that holds variable size pens perpendicular to the work plane with an adjustable handle. The objectives were the design of an AT that stabilizes the user's motion during handwriting and drawing while simplifying the concept to render the device affordable and accessible. In the short term, future work includes the manufacturing of the device and clinical validation with potential users.

**ACKNOWLEDGEMENTS**

This work is supported by the *Fonds de recherche du Québec – Nature et technologies* (FRQNT) and Dr. Campeau-Lecours's startup funds at CIRRIS and Université Laval.

**REFERENCES**

[1] Missiuna, C., Rivard, L., & Pollock, N. (2004). They're Bright but Can't Write: Developmental Coordination Disorder in School Aged Children. *Teaching Exceptional Children Plus, 1*(1), n1.

[2] Smoker, T. J., Murphy, C. E., & Rockwell, A. K. (2009). *Comparing memory for handwriting versus typing.* Paper presented at the Proceedings of the Human Factors and Ergonomics Society Annual Meeting.

[3] Longcamp, M., Zerbato-Poudou, M.-T., & Velay, J.-L. (2005). The influence of writing practice on letter recognition in preschool children: A comparison between handwriting and typing. *Acta psychologica,119*(1),67-79.

[4] Mueller, P. A., & Oppenheimer, D. M. (2014). The pen is mightier than the keyboard: Advantages of longhand over laptop note taking. *Psychological science, 25*(6), 1159-1168.

[5] Pontart, V., Bidet-Ildei, C., Lambert, E., Morisset, P., Flouret, L., & Alamargot, D. (2013). Influence of handwriting skills during spelling in primary and lower secondary grades. *Frontiers in psychology, 4*, 818.

[6] Idris, N. (2009). Enhancing students' understanding in calculus trough writing. *International Electronic Journal of Mathematics Education, 4*(1), 36-55.

[7] James, K. H., & Atwood, T. P. (2009). The role of sensorimotor learning in the perception of letter-like forms: Tracking the causes of neural specialization for letters. *Cognitive Neuropsychology, 26*(1), 91-110.

[8] Wu, F.-G., Chang, E., Chen, R., & Chen, C.-H. (2003). Assistive drawing device design for cerebral palsy children. *Technology and Disability, 15*(4), 239-246.

[9] Pedemonte, N., Laliberté, T., & Gosselin, C. (2013). *A bidirectional haptic device for the training and assessment of handwriting capabilities.* Paper presented at the World Haptics Conference (WHC), 2013.

[10] Shire, K. A., Hill, L. J., Snapp-Childs, W., Bingham, G. P., Kountouriotis, G. K., Barber, S., & Mon-Williams, M. (2016). Robot Guided 'Pen Skill'Training in Children with Motor Difficulties. *PloS one, 11*(3), e0151354.

[11] Henderson, S., Skelton, H., & Rosenbaum, P. (2008). Assistive devices for children with functional impairments: impact on child and caregiver function. *Developmental Medicine & Child Neurology, 50*(2), 89-98.

[12] Brown, T. (2009). Change by design, 1-5

[13] Campeau-Lecours, A., Lamontagne, H., Latour, S., Fauteux, P., Maheu, V., Boucher, F., ... & L'Ecuyer, L. J. C. (2017). Kinova Modular Robot Arms for Service Robotics Applications. *International Journal of Robotics Applications and Technologies (IJRAT)*, *5*(2), 49-71.

[14] Lebrasseur, A., Lettre, J., Routhier, F., Archambault, P., & Campeau-Lecours, A. (2018, July). Assistive robotic device: evaluation of intelligent algorithms. RESNA.

[15] Turgeon, P., Laliberté, T., Routhier, F., Campeau-Lecours, A. (2019). Preliminary design of an active stabilization assistive eating device for people living with movement disorders. *IEEE/RAS-EMBS International Conference on Rehabilitation Robotics (ICORR 2019)*, Toronto, Canada.

[16] Turgeon, P., Dubé, M., Laliberté, T., Routhier, F., Archambault, P., Campeau-Lecours, A. (2019). Mechanical design of a new assistive eating device for people living with spasticity. *Assistive Technology Journal.*